\documentclass{article}

\usepackage[preprint]{neurips_2023}

\usepackage{natbib}
\setcitestyle{numbers,square}
\usepackage[utf8]{inputenc} 
\usepackage[T1]{fontenc}    
\usepackage{hyperref}       
\usepackage{url}            
\usepackage{booktabs}       
\usepackage{amsfonts}       
\usepackage{nicefrac}       
\usepackage{microtype}      
\usepackage{xcolor}         
\usepackage{amsmath}
\usepackage{amssymb}
\usepackage{enumitem}
\usepackage{array}
\usepackage{graphicx}
\usepackage{color}
\usepackage{multicol}
\usepackage{appendix}
\usepackage{multirow}

\usepackage{sectsty}
\usepackage{chngcntr}
\usepackage{remreset}
\usepackage{makecell}
\usepackage{soul}
\usepackage{caption} 
\usepackage{subcaption}
\usepackage{accents}
\usepackage[nameinlink,capitalise]{cleveref}
\newcolumntype{C}[1]{>{\centering\arraybackslash}p{#1}}
\newcolumntype{L}[1]{>{\arraybackslash}p{#1}}

\title{ChatFace: Chat-Guided Real Face Editing via Diffusion Latent Space Manipulation}

\author{
    Dongxu Yue  \\
    Peking University   \\
    \texttt{yuedongxu@stu.pku.edu.cn}   \\
    \And
    Qin Guo   \\
    Peking University   \\
    \texttt{guoqin@stu.pku.edu.cn} \\
    \And
    Munan Ning   \\
    Peking University   \\
    \texttt{munanning@pku.edu.cn} \\
    \And
    Jiaxi Cui   \\
    Peking University   \\
    \texttt{jiaxicui446@gmail.com} \\
    \And
    Yuesheng Zhu\thanks{Corresponding author}  \\
    Peking University   \\
    \texttt{zhuys@pku.edu.cn}   \\
    \And
    Li Yuan$^*$  \\
    Peking University   \\
    \texttt{liyuan@pku.edu.cn}   \\
}

\begin{document}

\maketitle


\begin{figure}[htp]
 \centering
 \includegraphics[width=1.0\columnwidth]{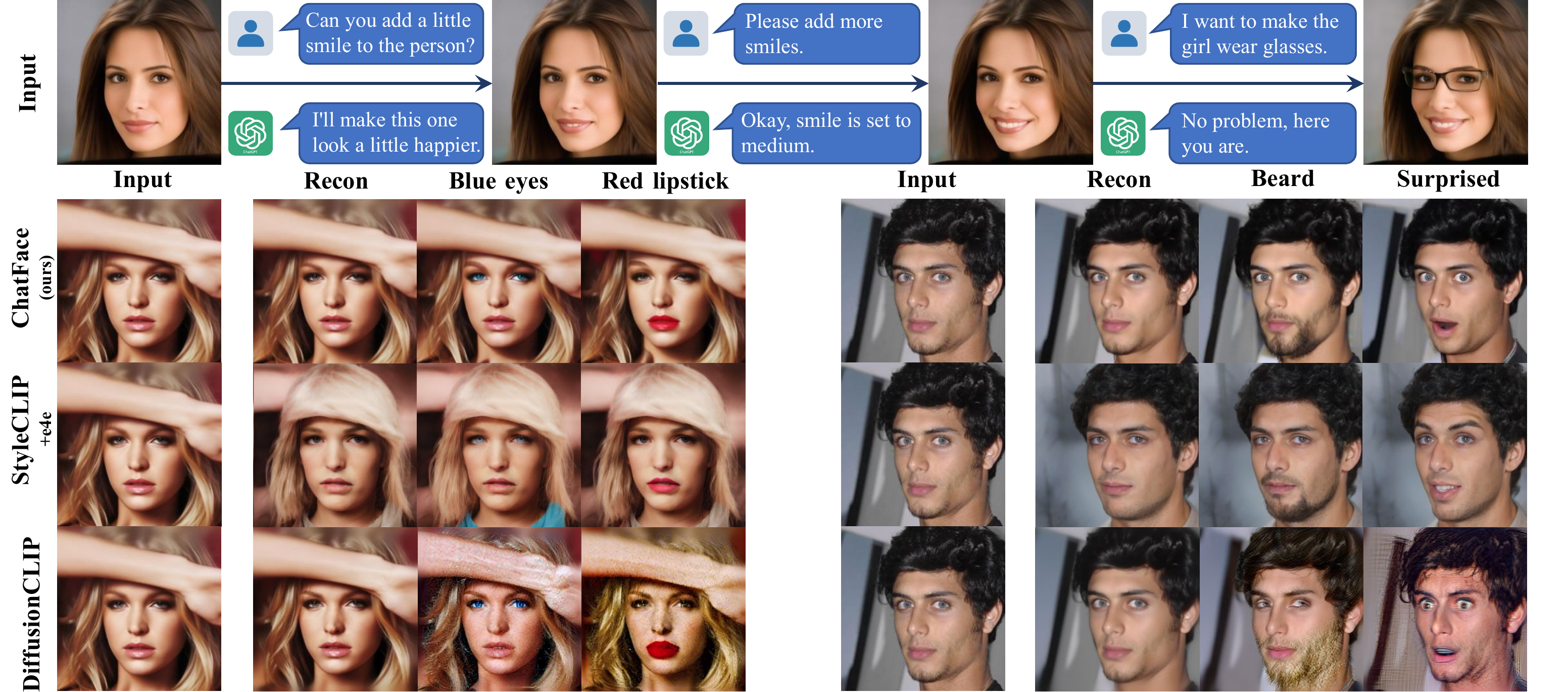}
 \caption{Examples of real facial image manipulations using ChatFace. Top row: our interactive system perform high-quality edits on a facial image provided by user. Bottom: we show manipulation results compared with other methods across multiple attributes.}
 \label{fig:motivation}
\end{figure}

\begin{abstract}  
Editing real facial images is a crucial task in computer vision with significant demand in various real-world applications. While GAN-based methods have showed potential in manipulating images especially when combined with CLIP, these methods are limited in their ability to reconstruct real images due to challenging GAN inversion capability. 
Despite the successful image reconstruction achieved by diffusion-based methods, there are still challenges in effectively manipulating fine-gained facial attributes with textual instructions.
    To address these issues and facilitate convenient manipulation of real facial images, we propose a novel approach that conduct text-driven image editing in the semantic latent space of diffusion model. By aligning the temporal feature of the diffusion model with the semantic condition at generative process, we introduce a stable manipulation strategy, which perform precise zero-shot manipulation effectively. Furthermore, we develop an interactive system named ChatFace, which combines the zero-shot reasoning ability of large language models to perform efficient manipulations in diffusion semantic latent space. This system enables users to perform complex multi-attribute manipulations through dialogue, opening up new possibilities for interactive image editing. Extensive experiments confirmed that our approach outperforms previous methods and enables precise editing of real facial images, making it a promising candidate for real-world applications. Project page: https://dongxuyue.github.io/chatface/

\end{abstract}

\section{Introduction}
Using natural language for image generation and manipulation is a straightforward and intuitive approach for humans. Since the emergence of Generative Adversarial Networks (GANs)\cite{goodfellow2020generative}, methods for image synthesis and editing have been extensively explored. 
Text-driven image editing has gained popularity through the incorporation of the supervision capability of CLIP\cite{radford2021learning} into approaches based on StyleGAN\cite{karras2019style, karras2020analyzing}, such as StyleCLIP\cite{patashnik2021styleclip}, which enables zero-shot image manipulation\cite{gal2022stylegan, wei2022hairclip}. 
However, the effectiveness of GAN-based methods for editing real images is limited by their reliance on GAN inversion to map real images into a semantic latent space. State-of-the-art encoder-based GAN inversion methods\cite{alaluf2021restyle, richardson2021encoding, tov2021designing} often fail to accurately reconstruct the original real images\cite{kim2022diffusionclip}, which in turn hinders their ability to edit real images, further restricting their real-world application. 
As illustrated in the bottom of Figure \ref{fig:motivation}, StyleCLIP with e4e\cite{tov2021designing} fails to reconstruct the girl's arms faithfully and results in noticeable changes to her facial identity. This problem becomes even more pronounced when dealing with real facial images that exhibit greater variations, leading to unintended change in the resulting images.


Recently, diffusion models\cite{sohl2015deep, ho2020denoising} have achieved impressive results in image generation, allowing for high-quality and diverse synthesis of images based on a text description\cite{rombach2022high, ramesh2022hierarchical, saharia2022photorealistic}. 
However, the application of diffusion models for semantic editing and manipulation of real images, especially when modifying local facial attributes, remains a challenge. 
Fortunately, Diffusion Autoencoders (DAE)\cite{preechakul2022diffusion}, based on denoising diffusion implicit models (DDIM) \cite{song2020denoising}, leverage an image encoder to explore a semantically rich space, leading to exceptional image inversion and reconstruction capabilities.
DAE also introduces an classifier to identify specific editing directions for some attributes. Nevertheless, all manipulations are constrained to pre-defined directions, significantly limiting users' creativity and imagination. Annotating additional data and retraining the classifier for new editing directions is necessary. 

To this end, one nature approach is to use CLIP to modify the latent code towards a given text prompt. However, we find this often results in unstable manipulations with unintended change. To address these limitations, we propose a new face editing pipeline which can perform arbitrary facial attribute manipulation in real images.
Specifically, we start with the input semantic code gained from aforementioned DAE and build a mapping network to yield the target code. Subsequently, we introduce a Stable Manipulation Strategy (SMS) to perform linear interpolation in diffusion semantic latent space by aligning the temporal feature of the diffusion model with the semantic condition at generative process, which enable precise zero-shot face manipulation of real images. 


Considering the widespread demand for editing real facial images, we aim to build an user-friendly system in an interactive manner, that can fulfill users' editing intentions effectively. The emergence of large language models (LLMS)\cite{chung2022scaling, brown2020language, touvron2023llama}, such as ChatGPT, has provided a new approach to addressing this problem, given their impressive language comprehension, generation, interaction, and reasoning capabilities.  
Moreover, the integration of LLMs with existing image models has been investigated recently\cite{lu2023multimodal, li2023videochat}. 

In this work, we present ChatFace, an advanced multimodal system for editing real facial images based on the semantic space of diffusion models. LLMs parse the complex editing queries based on our designed editing specifications, and then $z_{edit}$ is activated in the semantic latent space of diffusion models through the dynamic combination of our trained mapping network. We improve the editing stability in training and ensures semantic information coherence across different information levels during the generation process of the diffusion model by SMS that mentioned above.
The contributions of our work can be summarized as follows:
\begin{itemize}[leftmargin=*]
\item We introduce ChatFace, which enables users to interactively perform high-quality face manipulations on real images without the constraints of predefined directions or the problems associated with GAN inversion.
\item We propose a novel editing approach and SMS to perform stable manipulation within the semantic latent space of diffusion models in zero-shot manner. 
\item Both qualitative and quantitative experiments demonstrate that our method enables fine-grained semantic editing of real facial images, indicating that ChatFace has advantages in generating visually consistent results.
\end{itemize}

\section{Related Works}
\label{gen_inst}

\textbf{Image Manipulation. } 
Studies have explored the potential of generative models for image editing in various ways\cite{ling2021editgan, nam2018text, hertz2022prompt, brooks2022instructpix2pix, kwon2022diffusion}, such as style transfer\cite{zhu2017unpaired}, image translation\cite{saharia2022palette}, semantic manipulation\cite{abdal2019image2stylegan, abdal2020image2stylegan++, shen2020interpreting}, local edits\cite{hou2022feat, revanur2023coralstyleclip}, and we focus on discuss the semantic manipulation based methods here. StyleGAN\cite{karras2019style, karras2020analyzing} has become the preferred choice for previous studies due to its rich semantic latent space and disentanglement properties. Recently, diffusion models surpass GANs in high-quality image generation without using the less stable adversarial training.\cite{dhariwal2021diffusion}. Investigations\cite{abdal2021styleflow, zhang2023shiftddpms, kwon2022diffusion} explored the semantic latent space of diffusion models which can be utilized for image manipulation. Specifically, some works\cite{preechakul2022diffusion, shen2020interpreting, alaluf2021only} use annotated images as supervision to predict editing directions in the latent space, while others explore disentangled semantic manipulation directions in an unsupervised manner\cite{voynov2020unsupervised, wang2021geometry, yang2023disdiff, park2023unsupervised}. While these approaches yield great editing results, they are constrained by pre-defined directions for image manipulation. Recently, several text-to-image manipulation methods based on StyleGAN have been proposed\cite{patashnik2021styleclip, gal2022stylegan, xia2021tedigan, zhu2022one, wei2022hairclip, wang2022maniclip, kocasari2022stylemc}. These methods have to inverting real images to the latent space through GAN inversion, which makes faithful image reconstruction challenging. Text-driven image manipulation performance is further boosted in DiffusionCLIP\cite{kim2022diffusionclip} and Asyrp\cite{kwon2022diffusion}, where DDIM acts as encoder to enables faithful image inversion and reconstruction. However, due to the lack of a disentangled semantic latent space, they have difficulties in editing facial images without affecting other unintended attributes. In Section\ref{sec:4}, we demonstrate that our proposed method offers more effective manipulation of real facial images based on text inputs.

\textbf{Interactive Image Editing Systems. } 
An ideal interactive system for editing real facial images should be able to engage in a dialogue with the user based on their editing queries.  One recent relevant work in this domain is Talk-to-edit\cite{jiang2021talk}, which employs a text encoder to analyze the user's input, associating it with pre-defined facial attributes, and subsequently generates edited latent codes into the image domain through a generative adversarial network. Although attempts have been made to enhance interactivity, it faces two main challenges. First, limited parsing capability of the text encoder, making it difficult to analyze and map complex user requests to multiple editing directions while accurately controlling the editing strength. Second, as previously mentioned, the encoder based GAN inversion capability is limited, particularly when it comes to editing real images with complex backgrounds. In contrast, ChatFace brings interactive editing into real-world applications with remarkable abilities in understanding and parsing complex user requests and accurate semantic editing control.

\textbf{LLMs in Vision. } 
Integrating Large Language Models (LLMs) into visual tasks holds great promise and has gained significant attention from researchers. Numerous studies\cite{wang2023chatvideo, li2023videochat} have investigated the Combination of ChatGPT with existing visual models, leading to the development of novel applications. Visual ChatGPT \cite{wu2023visual}maps user inputs to different functionalities of the image-based model, while HuggingGPT\cite{shen2023hugginggpt} further expands this by integrating ChatGPT with a wide range of AI models from Hugging Face. Furthermore, a recent work\cite{lu2023multimodal} proposes a method of infusing visual knowledge into LLMs by utilizing existing images as enhanced visual features for language models and expressing image descriptions in a multimodal manner. In this paper, we present the first attempt at applying LLM to editing real facial image via diffusion semantic latent space interactively. 

\section{Method}
\begin{figure}[htp]
 \centering
 \includegraphics[width=1.0\columnwidth]{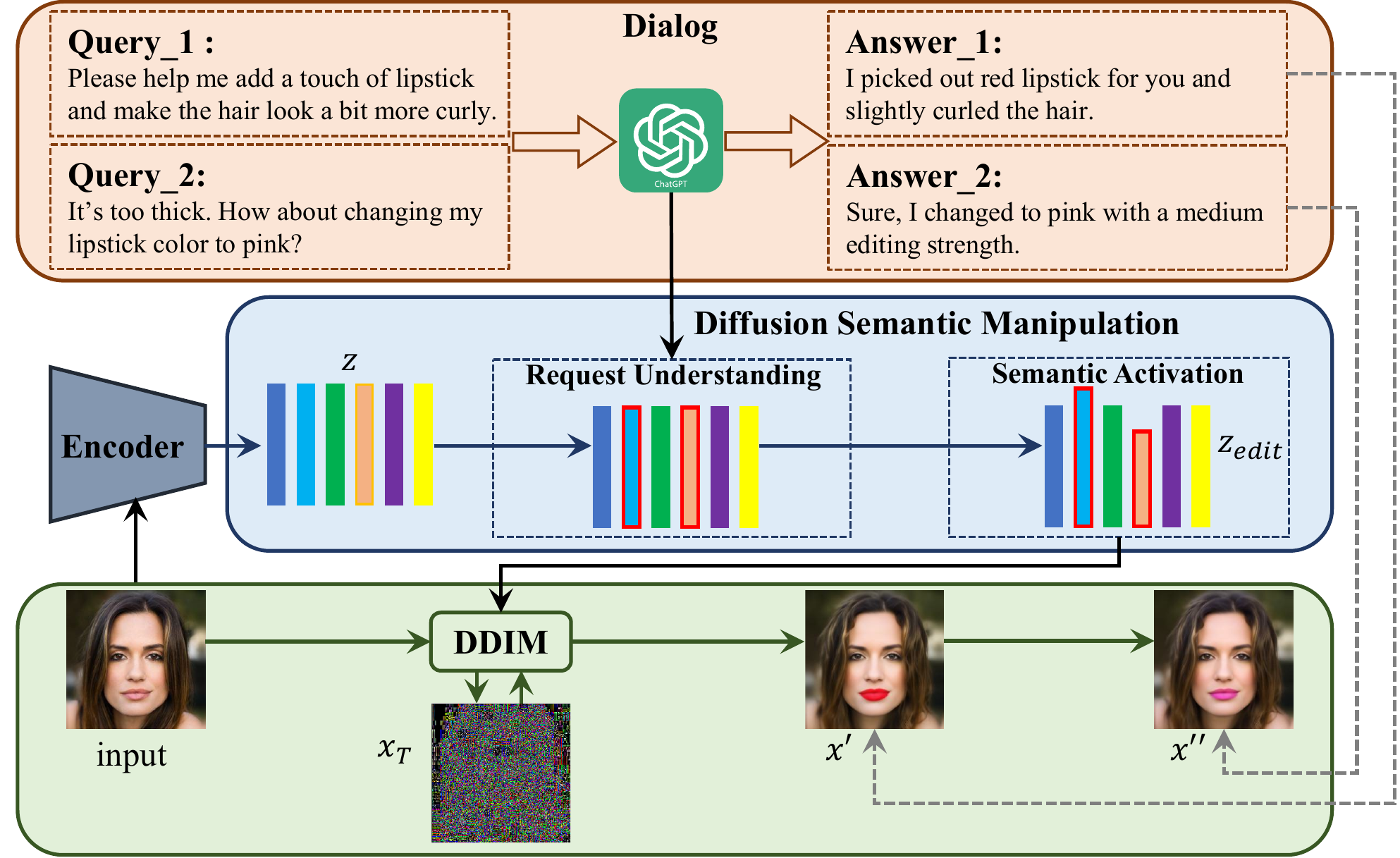}
 \caption{Overview of ChatFace inference pipeline. Large language model parsing queries from user for solving facial image editing tasks, which then enable the activation of corresponding facial attributes and control over the editing strength in diffusion semantic latent space.}
 \label{fig:pipline}
\end{figure}

The pipeline of our proposed ChatFace for real facial image manipulation is depicted in Figure\ref{fig:pipline}. Our objective is to develop a multimodal system for realistic facial image editing that allows users to edit their photos in an interactive manner. ChatFace consists of a large language model (LLM) as user request interpreter and controller, and a diffusion model with semantic latent space as a generator. 
By leveraging the LLM's capability to analyze diverse editing intentions, we manipulate the semantic latent space of the diffusion model with our stable manipulation strategy to achieve precise and fine-grained editing of real images. 
This interactive system empowers users to iteratively and continuously refine their edits until attaining the desired results. In the following, we provide a concise overview of the diffusion probability model and diffusion autoencoders, followed by a detailed explanation of our proposed method for semantic manipulation. Finally, we elucidate how ChatFace interacts with users to facilitate real facial image editing.

\subsection{Preliminaries}
\textbf{Diffusion Probabilistic Model.} 
The Denoising Diffusion Probabilistic Model (DDPM) is one of the most powerful generative models that consists of a forward process and a denoising backward process. The forward process is a Markov process where noise is gradually added to the data $x_0$ within time steps $1...T$, resulting in a series of corresponding latent variables denoted as $x_1...x_T$. Each step of the forward process follows a state transition equation: $q\left({x}_t \mid {x}_{t-1}\right):=\mathcal{N}\left(\sqrt{1-\beta_t} {x}_{t-1}, \beta_t \mathbf{I}\right)$ , where $\beta_t$ is a hyperparameter controlling the magnitude of variance.
In the reverse diffusion process, the transition from time step $t$ to $t-1$ can be interpreted as sampling from the distribution $p\left(x_{t-1} \mid x_t\right)$. This distribution can be further expanded as: $\mathcal{N}\left({x}_{t-1} ; {\mu}_\theta\left({x}_t, t\right), \sigma_\theta\left({x}_t, t\right) \mathbf{I}\right)$, where $\mu_\theta\left({x}_t, t\right)$ is a linear combination of a noise term ${\epsilon}_\theta\left({x}_t, t\right)$ predicted by a network and the noisy image $x_t$ at time step $t$. The model is trained with the $L_2$ loss between the predicted noise and the actual noise $\left\|\epsilon_\theta\left(x_t, t\right)-\epsilon\right\|_2^2$. Furthermore, denoising diffusion implicit models\cite{song2020denoising} (DDIM) has been proposed as a class of deterministic generative models. Similar to DPMs, DDIM gradually degrades the image $x_0$ to approximately Gaussian noise $x_T$ through a T-step forward process. In the reverse process, $x_{t-1}$ is obtained through the following denoising process:
\begin{equation}
    \boldsymbol{x}_{t-1}=\frac{1}{\sqrt{1-\beta_t}}\left(\boldsymbol{x}_t-\frac{\beta_t}{\sqrt{1-\alpha_t}} \boldsymbol{\epsilon}_\theta\left(\boldsymbol{x}_t, t\right)\right). 
\end{equation}
Through the deterministic generation process of DDIM, the image $x_0$ can also be encoded into a noise latent space $x_T$ as follows\cite{dhariwal2021diffusion}:
\begin{equation}
    x_{t+1}=\sqrt{\alpha_{t+1}} \frac{\boldsymbol{x}_t-\sqrt{1-\bar{\alpha}_t} \boldsymbol{\epsilon}_\theta\left(\boldsymbol{x}_t, t\right)}{\sqrt{\bar{\alpha}_t}}+\sqrt{1-\alpha_{t+1}} \epsilon_\theta\left(x_t, t\right).
\end{equation}

 However, studies\cite{kim2022diffusionclip} have shown that $x_T$ lacks semantic information of the input image, despite its remarkable reconstruction capabilities. 

\textbf{Diffusion Autoencoders.} 
In pursuit of  a semantically rich latent space, DAE\cite{preechakul2022diffusion} introduces an additional encoder to encode the input image $x$ into $Z$ space.
The encoding process is denoted as $z=Encoder(x)$, where $z$ is a high-dimensional vector in $\mathbb{R}^{512}$ that contains high-level semantic information of the image. Subsequently, taking $z$ as a conditioning variable, DDIM serves as a conditional stochastic encoder to generate the noise latent code $x_T$ as follows: 
\begin{equation}\label{eq:eq3}
\mathbf{x}_{t+1}=\sqrt{\alpha_{t+1}} \mathbf{f}_\theta\left(\mathbf{x}_t, t, \mathbf{z}\right)+\sqrt{1-\alpha_{t+1}} \epsilon_\theta\left(\mathbf{x}_t, t, \mathbf{z}\right), 
\end{equation}
where $\epsilon_\theta\left(\mathbf{x}_t, t, \mathbf{z}\right)$ is a noise predicted by a U-Net\cite{ronneberger2015u} with condition $z$, and $\mathbf{f}_\theta$ is defined as: 
\begin{equation}
    \mathbf{f}_\theta\left(\mathbf{x}_t, t, \mathbf{z}\right)=\frac{1}{\sqrt{\alpha_t}}\left(\mathbf{x}_t-\sqrt{1-\alpha_t} \epsilon_\theta\left(\mathbf{x}_t, t, \mathbf{z}\right)\right). \label{eq:eq4}
\end{equation}
After $T$ encoding steps, $x_T$ resides in $R^{H{\times}W{\times}3}$, which includes supplementary information from z, and it is possible to achieve precise reconstruction of real images when condition on $(x_T, z)$.

\begin{figure}[htp]
 \centering
 \includegraphics[width=1.0\columnwidth]{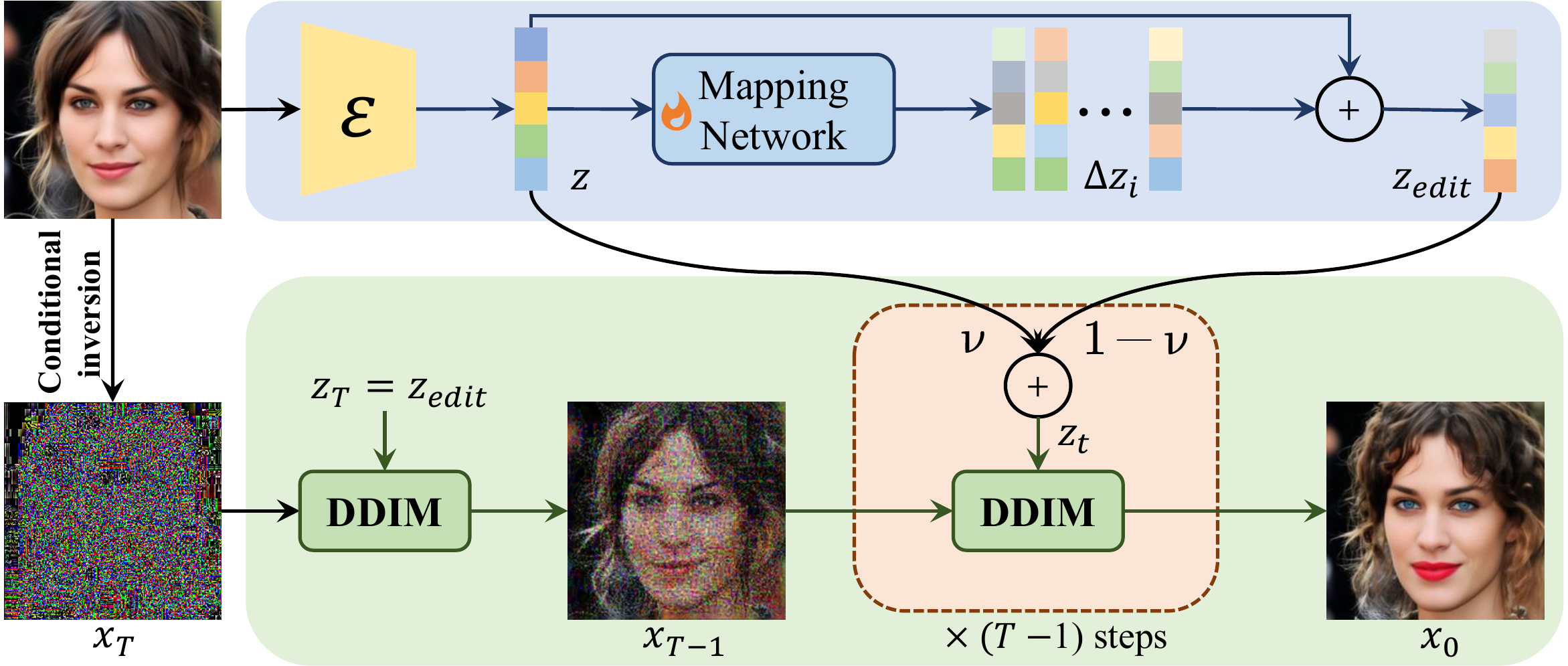}
 \caption{Our method consists of two parts: training a residual mapping network to obtain $z_{edit}$ in diffusion semantic latent space, and generating edited images with stable manipulation strategy.}
 \label{fig:training}
\end{figure}

\subsection{Semantic Manipulation}
\textbf{Architecture.}
The pipeline of our semantic manipulation method on diffusion model is illustrated in Figure\ref{fig:training}. 
The given input image $x$ is first encoded into the semantic latent space, denoted as $z$, where $z \in \mathbb{R}^{512}$. Subsequently, through the inversion process conditioned on $z$ via Eq\ref{eq:eq3}, the noise latent code $x_T$ is derived, which contains the low-level, randomly semantic information of the image\cite{preechakul2022diffusion}.
Our objective is to enable users to edit arbitrary attributes of real images according to their imagination. Given the significant distribution variations of real images within the semantic latent space, directly applying pre-defined editing directions to input images is challenging. Therefore, we trained a residual mapping network which is a lightweight MLP to infer manipulation directions $\Delta z$ given different input $z$, and then we inject the semantic editing offset as follows:
\begin{equation}
    z_{edit}=z+s*Mapping(z),
\end{equation}
where $s$ is a scale parameter controlling edit strength. During training phase, the value of $s$ is set to 1, while in the inference phase, this parameter is employed to regulate the degree of editing according to user's requests.

\textbf{Stable Manipulation Strategy.}
It has been shown that the generation process of the diffusion model from noise $x_T$ to generated image $x_{gen}$ is not uniform\cite{kwon2022diffusion}. In the initial denoising steps, it captures high-level features such as structures and shapes, whereas in the later steps, generate low-level features such as colors and textures. As mentioned, the semantic space $Z$ contains rich high-level information of the input image. However, when the same semantic condition $z_{edit}$ is applied to all denoising steps, it can alter the desired attributes but may lead to the loss of high-frequency details from the original image, resulting in unstable manipulation results. 
To address this problem, we propose an interpolation strategy that aligns the temporal features of the diffusion model with the semantic condition $z_t$ at each time step, as illustrated in the bottom of Figure\ref{fig:training}. Specifically, we obtain $z$ from the input image and compute $z_{edit}$ using the aforementioned residual mapping network. Then, we perform linear interpolation on a series of $z_t$ values between $z_{edit}$ and $z_0$ as follows: 
\begin{equation}
    z_t=Lerp(z_{edit},z;\nu), 
\end{equation}
where $\nu=t/T$, $t \in [0, 1, 2...T]$, and $T$ is the number of time steps for generation. 
Subsequently, taking $z_{t}$ as a factor on conditional DDIM and run generative process for $T$ steps, we can generate the edited image $x_{edit}$ that possess the desired visual attributes while preserving unrelated attributes: 
\begin{equation}
    x_{t-1}=\sqrt{\alpha_{t-1}}\mathbf{f}_\theta\left(\mathbf{x}_t, t, \mathbf{z_{t}}\right)+\sqrt{1-\alpha_{t-1}} \frac{\mathbf{x}_t-\sqrt{\alpha_t} \mathbf{f}_\theta\left(\mathbf{x}_t, t, \mathbf{z_{t}}\right)}{\sqrt{1-\alpha_t}}, 
\end{equation}
where $\mathbf{f}_\theta$ is define in Equation \ref{eq:eq4}.

\textbf{Training Objectives.}
 To achieve fine-grained editing of arbitrary facial attributes in real images, we have developed three types of losses to impose constraints on different objectives. Specifically, given an input image $x_0$, the corresponding semantic latent code $z_0$, and $z_{edit}$,  
 we introduce a reconstruction loss in image domian and $L_2$ norm in latent space to preserve unrelated semantics as follows:
 \begin{equation}
     \mathcal{L}_{pre}=\left\|\boldsymbol{x}_0-D(z_{edit})\right\|_1 + \left\|{\Delta}z\right\|_2, 
 \end{equation}
where $D(.)$ represents the DDIM decoder that applies our Stable Manipulation Strategy (SMS) which generates image from $x_T$, and ${\Delta}z=z_{edit}-z_0$.
As our focus is on manipulating human portrait images while preserving their identity, we incorporate a face identity loss to maintain consistency throughout the editing process as:
\begin{equation}
    \mathcal{L}_{i d}=1-\cos \left\{R\left(D\left(z_0\right)\right), R(D(z_{edit}))\right\}, 
\end{equation}
where R(·) indicates the pretrained ArcFace network\cite{deng2019arcface}. Following StyleGAN-NADA\cite{gal2022stylegan}, we incorporate the CLIP direction loss, which measures the cosine distance between the edited image and the desired text prompt. 
\begin{equation}
    \mathcal{L}_{direction }\left(D(z_{edit}), y_{\mathrm{tar}} ; D\left(z_0\right), y_{\mathrm{ref}}\right):=1-\frac{\langle\Delta I, \Delta T\rangle}{\|\Delta I\|\|\Delta T\|},
\end{equation}
where $
\Delta T=E_T\left(y_{\mathrm{tar}}\right)-E_T\left(y_{\mathrm{ref}}\right), \Delta I=E_I\left(D(z_{edit})\right)-E_I\left(D\left(z_0\right)\right)
$, and $E_T$ ,$E_I$ denote the CLIP text encoder and image encoder respectively and $y_{\mathrm{tar}}$ and $y_{\mathrm{ref}}$ represent the target and reference text, respectively. Finally, the total loss can be written as: 
\begin{equation}
    \mathcal{L}_{total}=\lambda_{pre}\mathcal{L}_{pre}+\lambda_{id}\mathcal{L}_{id}+\lambda_{dir}\mathcal{L}_{direction}. 
\end{equation}
The weights for each loss, denoted as $\lambda_{pre}$, $\lambda_{id}$, and $\lambda_{dir}$. Specifically, we set $\lambda_{recon}=0.2$, $\lambda_{id}=0.5$, and $\lambda_{dir}=2.0$ in our following experiments. 


\subsection{Chat to Edit}
ChatFace is an interactive system that includes an LLM as user request interpreter and controller. Given an editing query $Q$, we design editing specifications for ChatGPT to parse and extract the interested facial attributes and corresponding editing strength from $Q$, and then map to semantic offset in the diffusion latent space. Finally, the system generates a response based on the extracted information, incorporating the desired modifications as specified by the user. 

\textbf{Editing Intention Understanding.}
We encourage large language models to understand and extract relevant attributes from $Q$, and decompose them into a series of structured attributes. To this end, we design a unified template for editing specifications, allowing LLM to parse the user's editing intent through slot filling. ChatFace employs three slots for editing intent parsing: desired editing attribute $A$, editing strength $S$, and diffusion sample step $T$, respectively. By injecting demonstrations into the prompts, ChatFace allows the large language model to better understand the editing intention, facilitating the analysis of the input queries and decompose them into combinations of ${A, S, T}$. In cases where users provide ambiguous queries, the LLM recognizes the most similar attributes and defaults to a moderate edit setting. 

\textbf{Semantic Activation.}
After parsing the queries, ChatFace needs to align the attributes with the manipulation directions in the semantic latent space of the diffusion model. For this purpose, we construct a database of attribute mapping network, which is  obtained through the training process described above and accompanied by detailed functional descriptions. Furthermore, we treat this issue as a multiple-choice problem, where the mapping network is presented as options given the context. Subsequently, we activate $z_{edit}$ for various attributes as follows: 
\begin{equation}
    z_{edit} = \sum s_i{\Delta}z_i+z_0, 
\end{equation}
where $s_i$ is the editing strength of the corresponding attribute extracted from the queries. 

\section{Experiments}\label{sec:4}
We evaluate the performance of our proposed method in real facial image editing tasks, and then we compare ChatFace with existing methods both qualitatively and quantitatively. We conducted ablation study to validate the effectiveness of our stable manipulation strategy and setting. Implementation details and additional experimental results and are provided in the Appendix\ref{sec:appA}. 

\begin{figure}[htp]
 \centering
 \includegraphics[width=1.0\columnwidth]{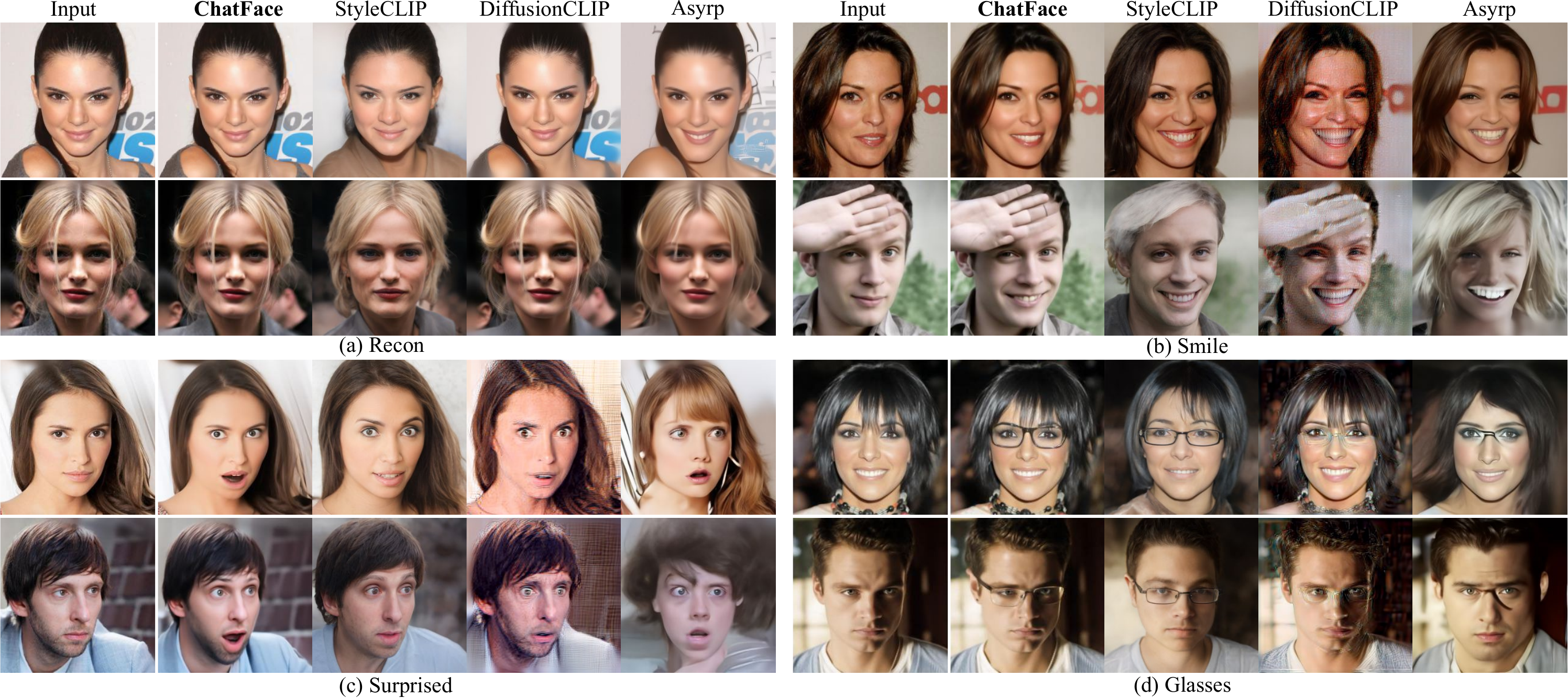}
 \caption{Comparison results with the state-of-the-art image manipulation methods: StyleCLIP\cite{patashnik2021styleclip}, DiffusionCLIP\cite{kim2022diffusionclip}, and Asyrp\cite{kwon2022diffusion}.}
 \label{fig:compare}
\end{figure}

\begin{figure}[htp]
 \centering
 \includegraphics[width=1.0\columnwidth]{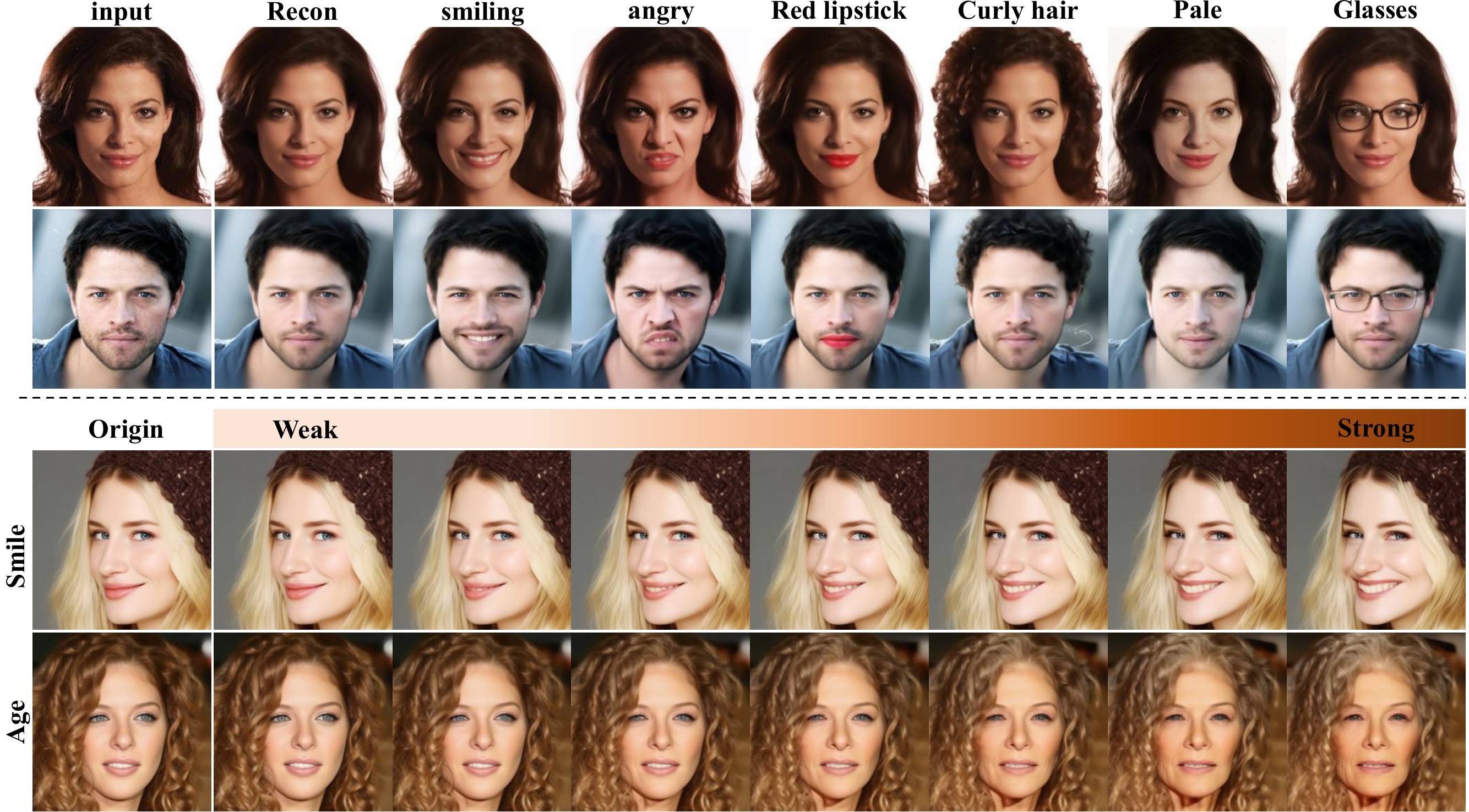}
 \caption{Our manipulation results on CelebA-HQ dataset with different semantics. The input images are shown in the first column and our results are shown in the corresponding row.}
 \label{fig:sinle_person}
\end{figure}

\subsection{Qualitative Evaluation}
\textbf{State-of-the-art Comparisons.}
Figure\ref{fig:compare} shows the visual comparison results, we compare our method with state-of-the-art text-guided image manipulation methods. We observe that StyleCLIP struggles to faithfully reconstruct real images, and local attribute modifications result in unintended change as described in Figure\ref{fig:compare}(a). For example, as shown in Figure\ref{fig:motivation} manipulating the "blue eyes" attribute also changes the girl's clothing color to blue. Furthermore, while DiffusionCLIP improves image reconstruction results of StyleCLIP, editing fine-grained facial attributes often affects the global visual features of the image as described in Figure\ref{fig:compare}(b) and (c). 
In contrast, our ChatFace perform efficient real image editing based on the input queries while preserving visual fidelity.

\begin{figure}[htp]
 \centering
 \includegraphics[width=1.0\columnwidth]{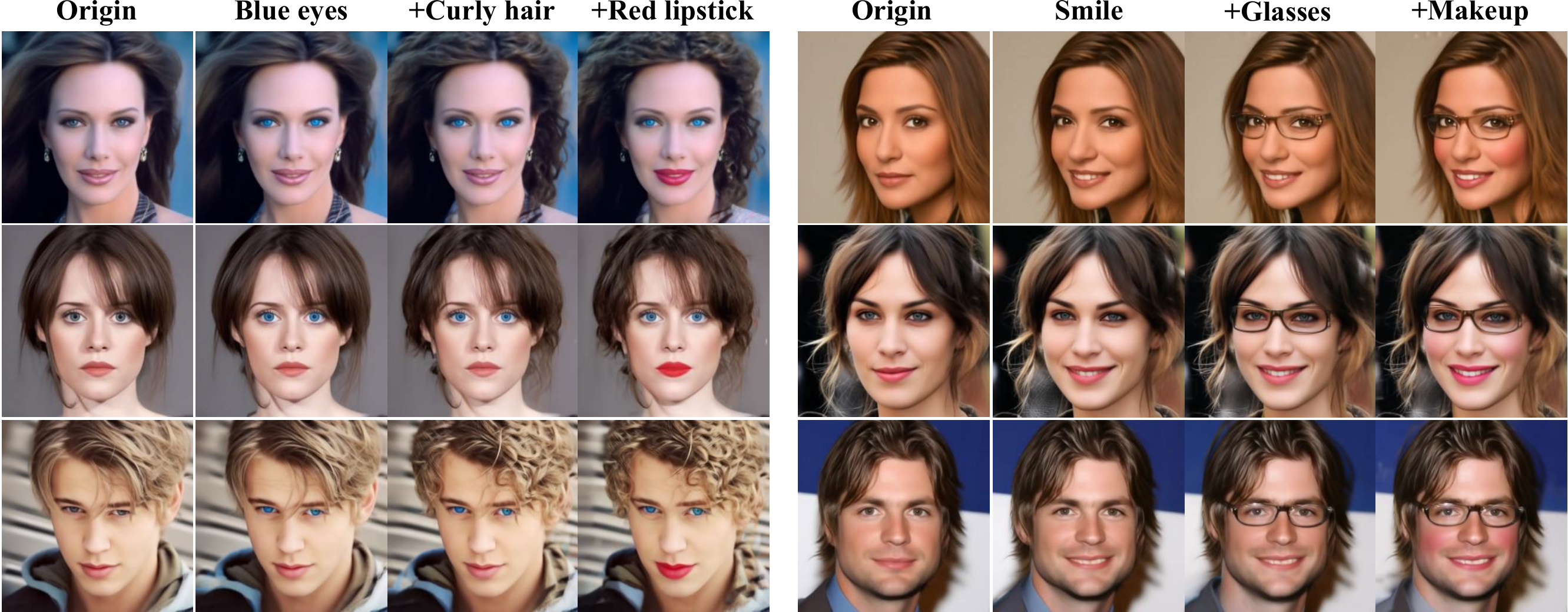}
 \caption{Manipulation results of multi-attribute transfer with queries consisting of gradually increasing attributes.}
 \label{fig:multi_attr}
\end{figure}

\textbf{Real Facial Image Editing.}
In Figure\ref{fig:sinle_person}, we demonstrate the effectiveness of our proposed ChatFace in performing various facial attribute edits, including expressions, local attributes, hairstyles, and global styles. Furthermore, we show a smooth morphing by scale strength parameter $s$ with SMS. We mainly focus on two aspects: first, consistent preservation of unrelated semantics in real images before and after manipulation, and second, maintaining a high correlation between the target attribute and the input editing queries. As observed, ChatFace successfully preserves the identity of the face and generates high-quality edited images. The diverse manipulation results showcase the robustness of our approach. Additional results can be found in the Appendix\ref{sec:appA2}.


\textbf{Multi-attribute Manipulation.}
We enable ChatFace to perform multi-attribute editing by sequentially incorporating the semantics of multiple attributes into real facial images, which are shown in Figure\ref{fig:multi_attr}. It's clear that ChatFace can generate progressive multi-attribute edits based on the user's queries, thereby demonstrating the continuous editing capability of our proposed method. 

\begin{table}[]
\caption{Quantitative evaluation and human evaluation results on CelebA-HQ\cite{karras2017progressive}. ChatFace achieves better performance in terms of $S_{dir}\uparrow$, SC$\uparrow$, ID$\uparrow$ and human evaluation score.}  
\label{tab:compare}
\centering
    \setlength\tabcolsep{2.0mm}
    \renewcommand\arraystretch{1.1}
    {
    \scalebox{1.0}{
        \begin{tabular}{c ccc cccc}
        \toprule[1pt]
        \textbf{}              & \multicolumn{3}{c}{Editing Performance} & \multicolumn{4}{c}{Human Evaluation}                             \\
        \cmidrule[1pt](lr){2-4}
        \cmidrule[1pt]{5-8}
                               & $S_{dir}\uparrow$    & SC$\uparrow$   & ID$\uparrow$   & Smile & Curly hair & Makeup & Glasses \\
        \midrule
        StyleCLIP~\cite{patashnik2021styleclip}     & 0.13                & 86.8\%             & 0.35             & 2.3 \%             & 5.1 \%                 & 1.6\%              & 3.5\%               \\
        Stylegan-NADA~\cite{gal2022stylegan} & 0.16                &  89.4\%             &  0.42             & 1.6\%               & 2.2\%                    & 2.2\%                & 0.9\%                 \\
        DiffusionCLIP~\cite{kim2022diffusionclip} & 0.18                & 88.1\%               & 0.76              & 0.9\%               &  6.7\%                   & 4.9\%               & 0.0\%                 \\
        Asyrp~\cite{kwon2022diffusion}         & 0.19                & 79.3\%              & 0.38               &  4.9\%              &   3.3\%                  &  0.9\%                &  1.4\%               \\
        \midrule
        ChatFace      & \textbf{0.21}                & \textbf{89.7\%}              &  \textbf{0.84}             & \textbf{90.3\%}               & \textbf{82.7\%}                    &\textbf{90.4\%}                 & \textbf{94.2\%}                 \\
        \bottomrule[1pt]
        \end{tabular}}
    }

\end{table}

\subsection{Quantitative Evaluation}
 Evaluating face image manipulation results is a challenging task. Nevertheless, following DiffusionCLIP\cite{kim2022diffusionclip}, we adopted three quantitative metrics to assess our proposed method. Directional CLIP similarity ($S_{dir}$) measures the similarity between the manipulated image and the corresponding text prompt using a pre-trained CLIP\cite{radford2021learning} model. Segmentation-consistency (SC), and face identity similarity (ID) are introduced to evaluate the semantic consistency and face identity between the results and the input images. As shown in the left part of Table\ref{tab:compare}, result indicates that ChatFace effectively manipulates real facial image attributes while maintaining consistency with the original images, outperform the compared methods on all metrics mentioned above.

\textbf{Human Evaluation.}
To evaluate the edited proformance of the compared methods, we conducted a user survey. We randomly collected 30 images from the CelebA-HQ dataset that were manipulated using four attributes (smile, curly hair, makeup, glasses). We used a survey platform to collect 5,000 votes from 45 participants with diverse backgrounds. First, participants were asked to choose the most semantically relevant results corresponding to the given attribute. Then, they were asked to evaluate the visual realism and identity consistency of the edited images to select the best image overall. The survey results are presented in the right part of Table\ref{tab:compare}, indicate that the majority of human subjects found our ChatFace model to have superior performance. We also assess the fluency of ChatFace in conversations and the accuracy of user intent extraction, further evaluation results are provided in the Appendix\ref{sec:appA3}.

\begin{figure}[htp]
 \centering
 \includegraphics[width=0.98\columnwidth]{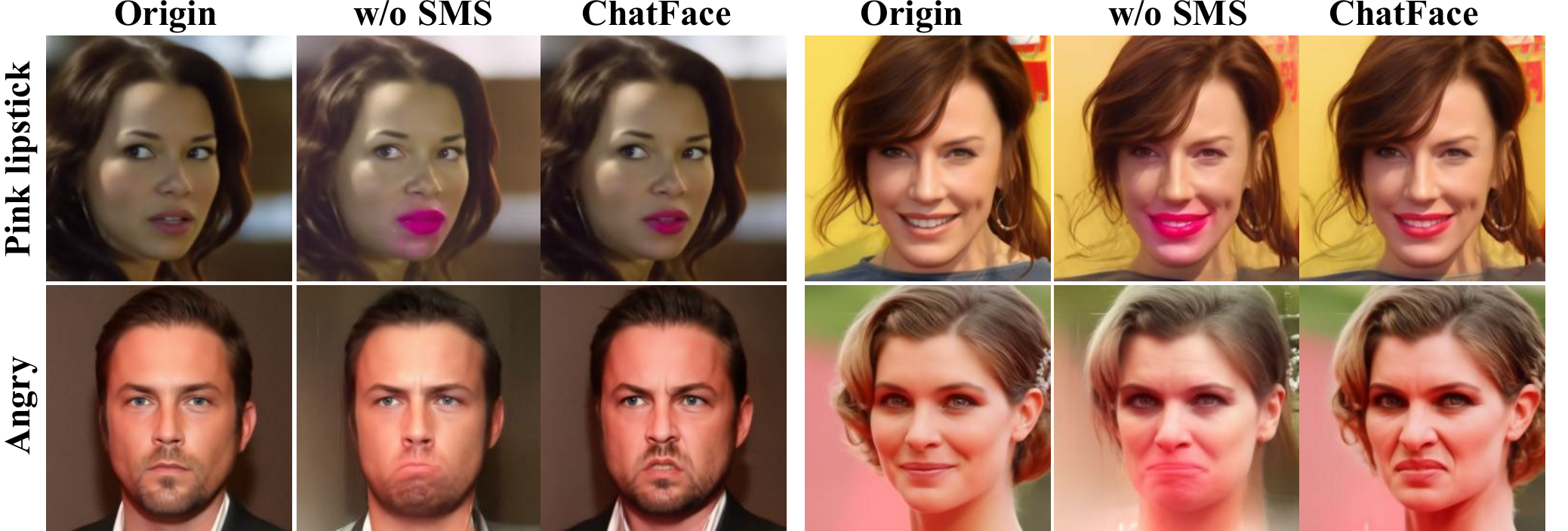}
 \caption{The effectiveness of our proposed stable manipulation strategy.}
 \label{fig:ablation_sms}
\end{figure}


\subsection{Ablation Analysis}
Our proposed Stable Manipulation Strategy (SMS) allows the semantic condition $z_t$ to match information levels across different generative process temporally, achieving more realistic manipulation results. To verify the effectiveness of our method, we demonstrate four samples manipulated by two different facial attributes. As shown in Figure\ref{fig:ablation_sms}, when editing the local attribute ``pink lipstick'', it can be observed that without using SMS, the pink color overflows from the lips, and the low-level semantic information of the original real image is not well preserved. Furthermore, when editing the facial expression of the character, ChatFace with SMS exhibits superior semantic consistency before and after the editing process. The additional analyses on ablation studies and hyperparameters are provided in Appendix\ref{sec:appC}.

\subsection{Case Study}
As a multimodal interactive system, ChatFace leverages a large language model to improve the semantic editing abilities of the diffusion model for manipulating real images by means of queries parsing and semantic activation. To demonstrate the effectiveness of ChatFace, we conducted a series of tests on a variety of editing tasks, and some selected cases are shown in Appendix\ref{sec:appB}. 

\section{Conclusions}
In this paper, we proposed ChatFace, a real facial image manipulation method within the semantic latent space of the diffusion model. We introduced a novel image manipulation method, which enable a wide variety of unique image manipulations with our stable manipulation strategy. We have also demonstrated that ChatFace provides fine-grained edit controls on complex editing tasks when combines large language model with the abilities of diffusion model, which enables semantically consistent and visually realistic facial editing of real images in an interactive manner. 

A limitation of our method is that it cannot be expected to manipulate images outside the domain of the pretrained DAE, and the generalization of our ChatFace in visually diverse datasets remains for further investigations. There are potential social risks on ChatFace, and we advise users to make use of our method for proper purposes. 

\clearpage
\bibliographystyle{plain}
\bibliography{ref}

\newpage


\begin{appendices}

\section{Details on Experiments}\label{sec:appA}
\subsection{Implementation Details}

We employ pre-trained Diffusion Autoencoders (DAE)\cite{kwon2022diffusion} with a resolution of 256 for image encoding and generation. The dimensions of the semantic code $z$ and the noise code $x_T$ are $\mathbb{R}^{512}$ and $\mathbb{R}^{256\times256\times3}$, respectively. To demonstrate the generalization and robustness of the ChatFace system, we trained our mapping network on the CelebA-HQ\cite{karras2017progressive}, while the DAE was trained on the FFHQ\cite{karras2019style}.  Our experiments employed 54 text prompts specifically designed for facial images, including expressions, hairstyles, age, gender, style, glasses, and more. The Ranger optimizer was used in our experiments\cite{Ranger}, and we set the learning rate to 0.2 and trained each attribute for 10,000 iterations with a batch size of 8. Our model was trained using 8 Nvidia 3090 GPUs, and we used $T=8$ for diffusion sample steps to generate edited images by default. For large language model, we utilized the GPT-3.5-turbo model, which can be accessed through OpenAI's API.

\textbf{Mapping network architecture.}
Our mapping network architecture is very simple and lightweight, consisting of only 4 layers of MLP. This enables us to efficiently combine and process complex tasks. The mapping network is trained to infer a manipulation direction in diffusion semantic latent space. We only need to train each text prompt once, and then we can perform semantic editing of the corresponding attribute on any real image. The architecture is specified in Table\ref{tab:architecture}.

\begin{table}[htbp]
\caption{Architecture of our mapping network.}  
\label{tab:architecture}
\centering
    \setlength\tabcolsep{2.0mm}
    \renewcommand\arraystretch{1.1}
    {
    \scalebox{1.0}{
    \begin{tabular}{c c}
       \toprule[1pt]
       Parameter & Setting \\
       \midrule
       Batch size & 8  \\
       MLP layers & 4  \\
       MLP hidden size & 512  \\
       $z$ size & 512  \\
       Learning rate & 0.2  \\
       Optimizer & Ranger  \\
       Train Diff $T$ & 8  \\
       Train $s$ & 1  \\
       \bottomrule[1pt]
    \end{tabular}
    }
    }
\end{table}

\textbf{User Request Understanding.}
The large language model takes a request from user and decomposes it into a sequence of structured facial attributes. We design a unified template for this task. Specifically, ChatFace designs three shots for editing intent parsing: desired editing attribute $A$, editing strength $S$, and diffusion sample step $T$. To this end, we inject demonstrations to ``teach'' LLM to understand the editing intention, and each demonstration consist of a user's request and the target facial attribute sequence, as shown in Table\ref{tab:demanstration}. We also show semantic activation details in the table.

\subsection{Additional Results}\label{sec:appA2}
In this section, we provide additional results to those presented in the paper. We begin with the manipulations a variety of images that are taken from CelebA-HQ, and then we perform manipulations on real images collected from the Internet.

\begin{figure}[htp]
 \centering
 \includegraphics[width=1.0\columnwidth]{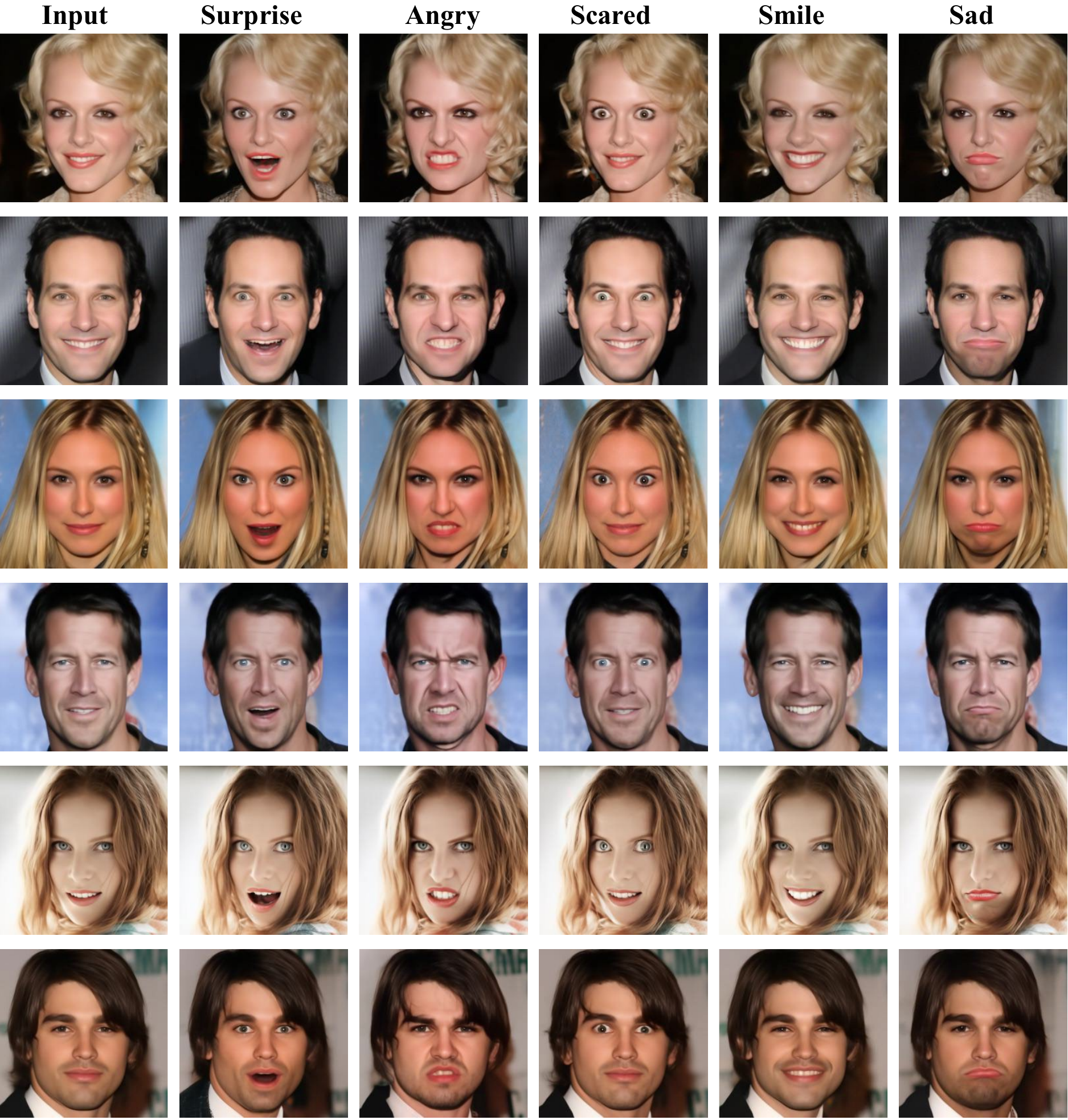}
 \caption{More visual results of expression edits.}
 \label{fig:app_expression}
\end{figure}

\begin{figure}[htp]
 \centering
 \includegraphics[width=1.0\columnwidth]{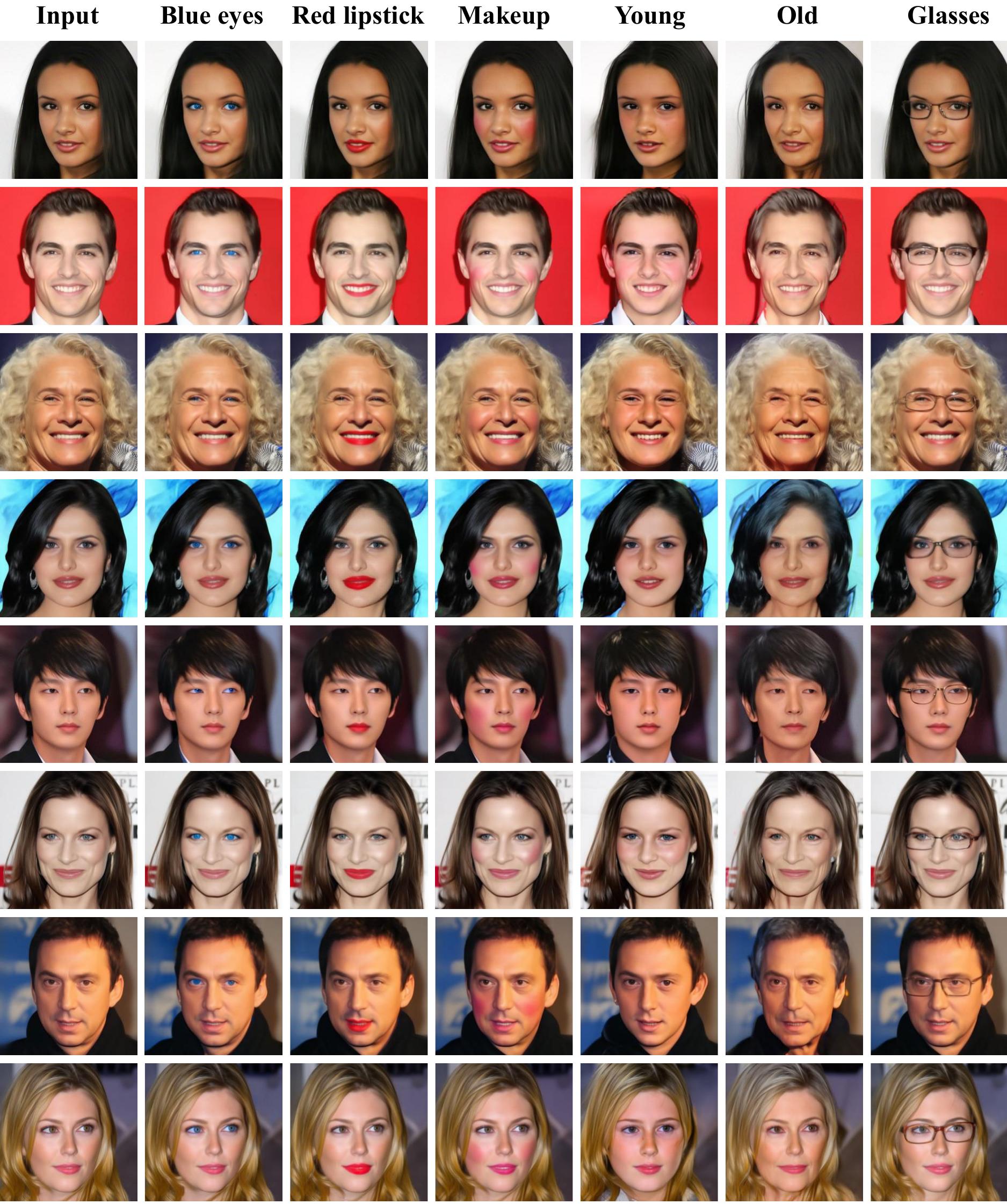}
 \caption{More visual results of local edits.}
 \label{fig:app_local}
\end{figure}

\begin{figure}[htp]
 \centering
 \includegraphics[width=1.0\columnwidth]{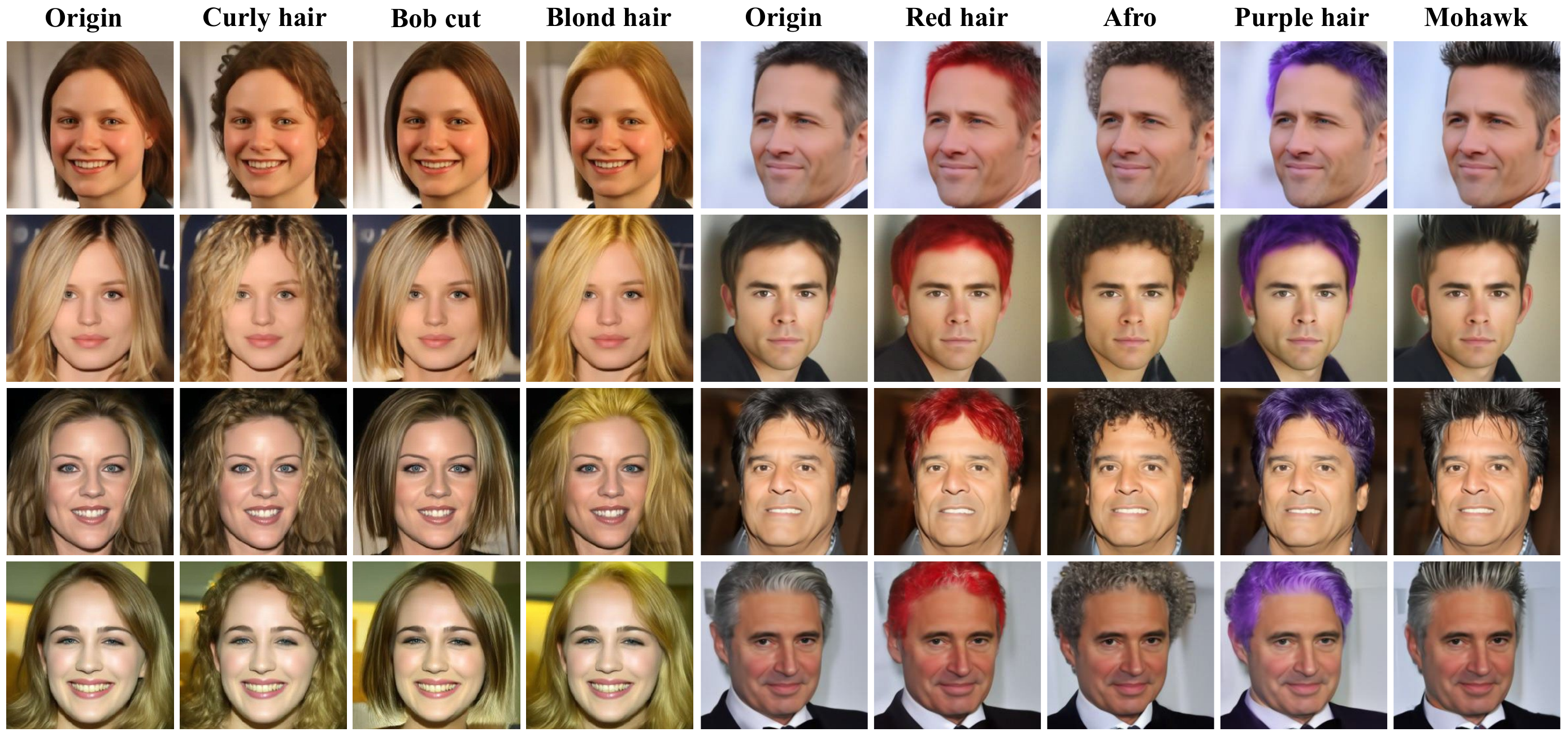}
 \caption{More visual results of hairstyle edits.}
 \label{fig:app_hairstyle}
\end{figure}

\begin{figure}[htp]
 \centering
 \includegraphics[width=1.0\columnwidth]{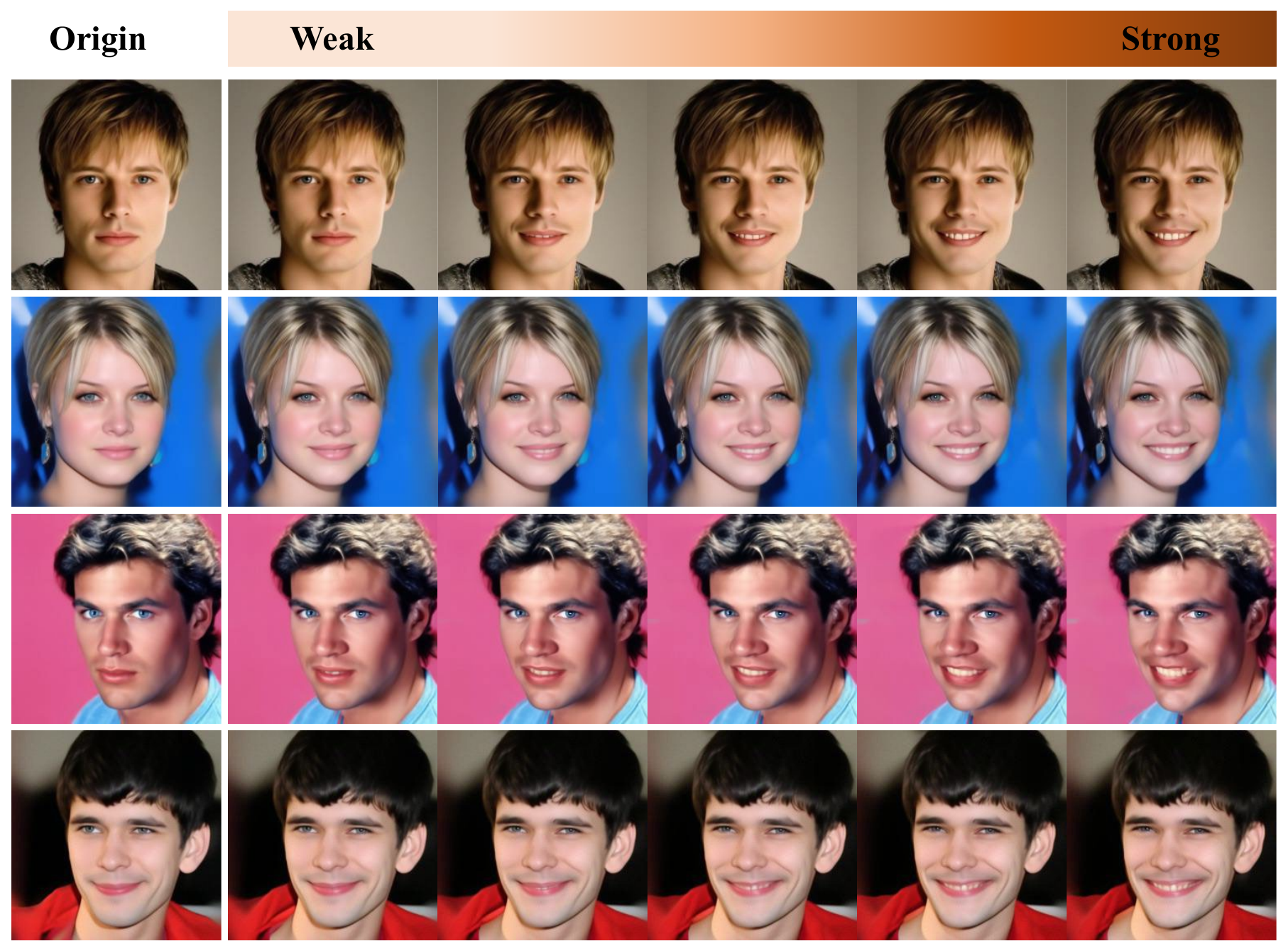}
 \caption{We demonstrate expression manipulation (driven by the prompt``a photo of a smile person'') for different manipulation strengths.}
 \label{fig:app_strength}
\end{figure}

\begin{figure}[htp]
 \centering
 \includegraphics[width=1.0\columnwidth]{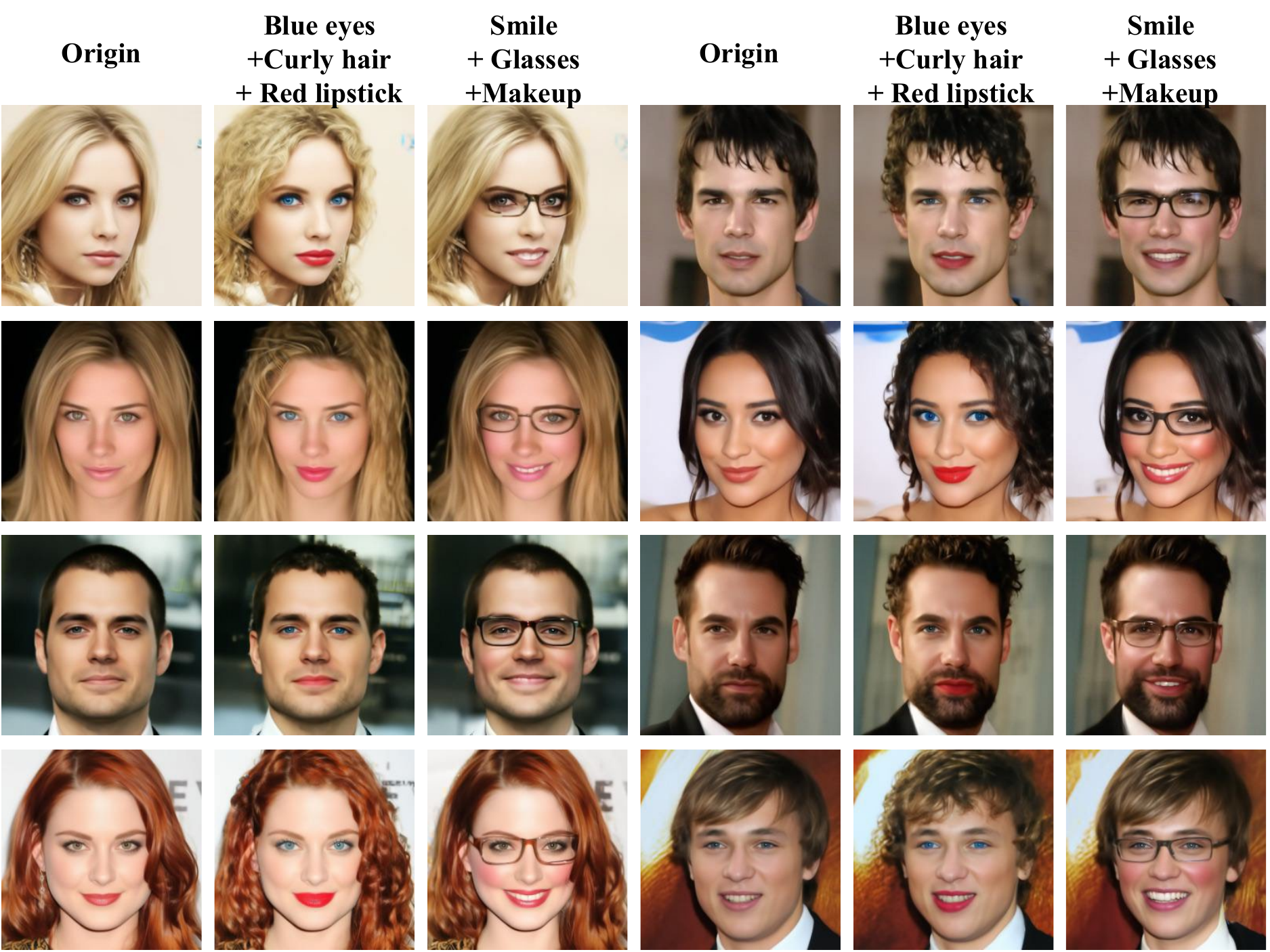}
 \caption{The visual results for multi-attribute manipulation.}
 \label{fig:app_multi}
\end{figure}

\textbf{Manipulation of images from CelebA-HQ.}
In Figure\ref{fig:app_expression} we show a verity of expression edits. In Figure\ref{fig:app_local} we show a large galley of local facial edits. Figure \ref{fig:app_hairstyle} shows hair style manipulations. We shows image manipulations driven by different editing strength which is derived from user's request in Figure \ref{fig:app_strength}. Figure\ref{fig:app_multi} demonstrates more results that ChatFace perform multi-attribute manipulations.

\begin{figure}[htp]
 \centering
 \includegraphics[width=1.0\columnwidth]{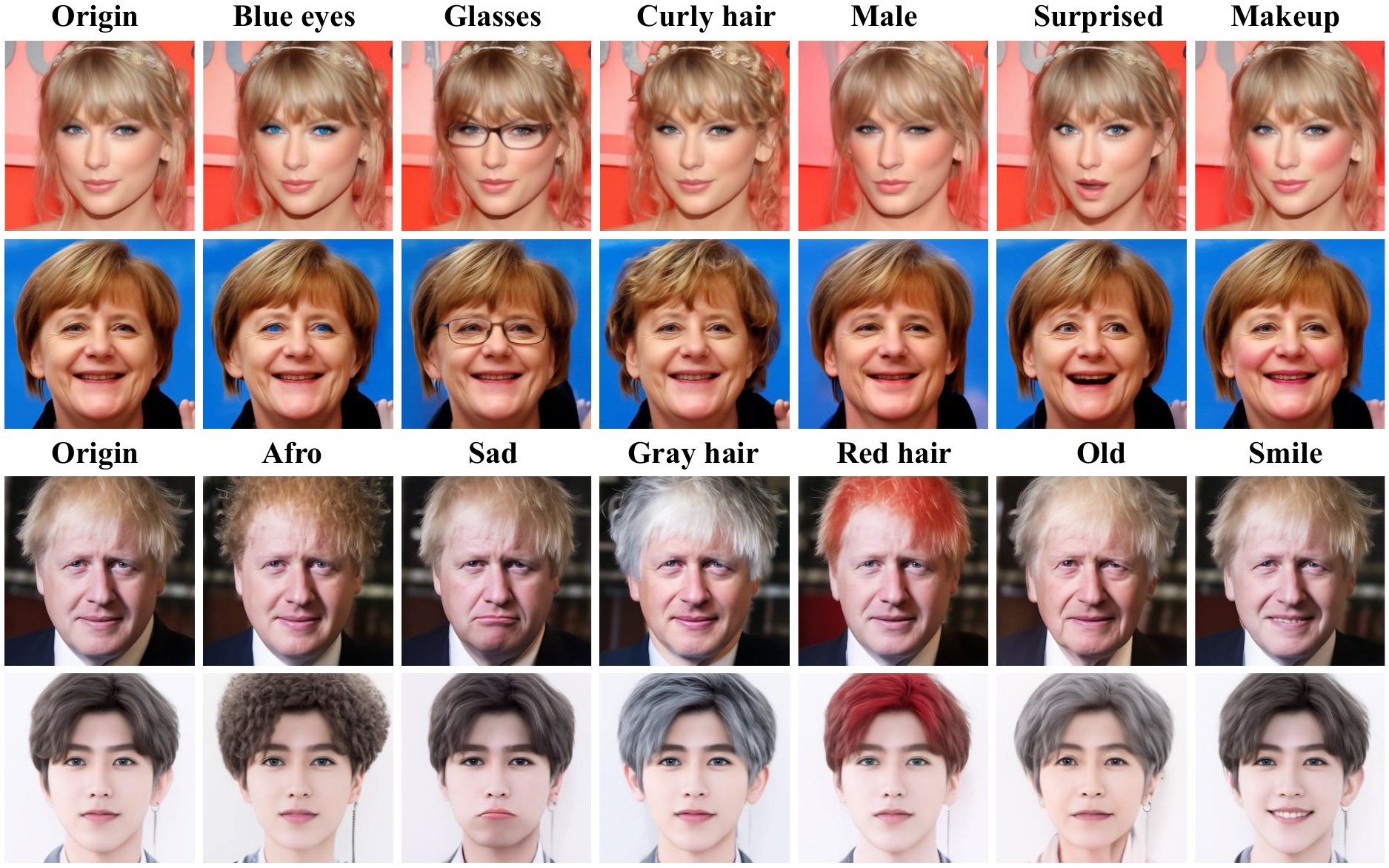}
 \caption{We perform manipulation on real facial images randomly collected from the Internet. The ChatFace demonstrates good generalization on these face images.}
 \label{fig:app_internet}
\end{figure}

\textbf{Manipulation of images from the Internet.}
We perform real face manipulations on images randomly collected from the Internet as shown in Figure\ref{fig:app_internet}. Our editing results look highly realistic and plausible.

\subsection{Human Evaluation of ChatFace}\label{sec:appA3}
In Section4 of the main text, we have demonstrated that ChatFace is capable of effectively manipulating real facial image attributes while maintaining consistency with the original images. We further evaluated the ability of  ChatFace to interpret user editing intentions and maintain conversational fluency during interactive usage through human subject evaluations. The results are presented in Table\ref{tab:app_user_study}.

\begin{table}[htbp]
\caption{Architecture of our mapping network.}  
\label{tab:app_user_study}
\centering
    \setlength\tabcolsep{2.0mm}
    \renewcommand\arraystretch{1.1}
    {
    \scalebox{1.0}{
    \begin{tabular}{c c c c}
       \toprule[1pt]
       Tasks & Good & Acceptable & Poor \\
       \midrule
       Request understanding & 39.6\% & 53.3\% & 7.1\%  \\
       Fluency of conversion & 67.6\% & 30.2\% & 2.2\%  \\
       \bottomrule[1pt]
    \end{tabular}
    }
    }
\end{table}



\subsection{More Results of Comparison}
Here, we provide details on the qualitative comparison of real facial image manipulation performance between our ChatFace and SOTA methods which are divided into GAN-based methods and diffusion-based methods. Specifically, We campare ChatFace with StyleCLIP-GD\cite{patashnik2021styleclip}, StyleGAN-NADA\cite{gal2022stylegan}, TediGAN\cite{xia2021tedigan}, DiffusionCLIP\cite{kim2022diffusionclip}, and Asyrp\cite{kwon2022diffusion}.

\textbf{Comparison setting.}
We followed the experimental setting as described in DiffusionCLIP\cite{kim2022diffusionclip}. For quantitative comparison, we use 1000 test images from CelebA-HQ, and we use the manipulation results for three attributes(makeup, tanned, gray hair). Please note that DiffusionClip and Asyrp are our reimplementation versions, and the comparative results are shown in Table\ref{tab:compare} of the main text. Following the settings in the paper of these methods, we use Encoder for Editing (e4e)\cite{tov2021designing}, ReStyle encoder\cite{alaluf2021restyle}, and pixel2style2pixel (pSp) encoder\cite{richardson2021encoding} respectively for the inversion of StyleCLIP, StyleGAN-NADA and TediGAN.

\begin{figure}[htp]
 \centering
 \includegraphics[width=1.0\columnwidth]{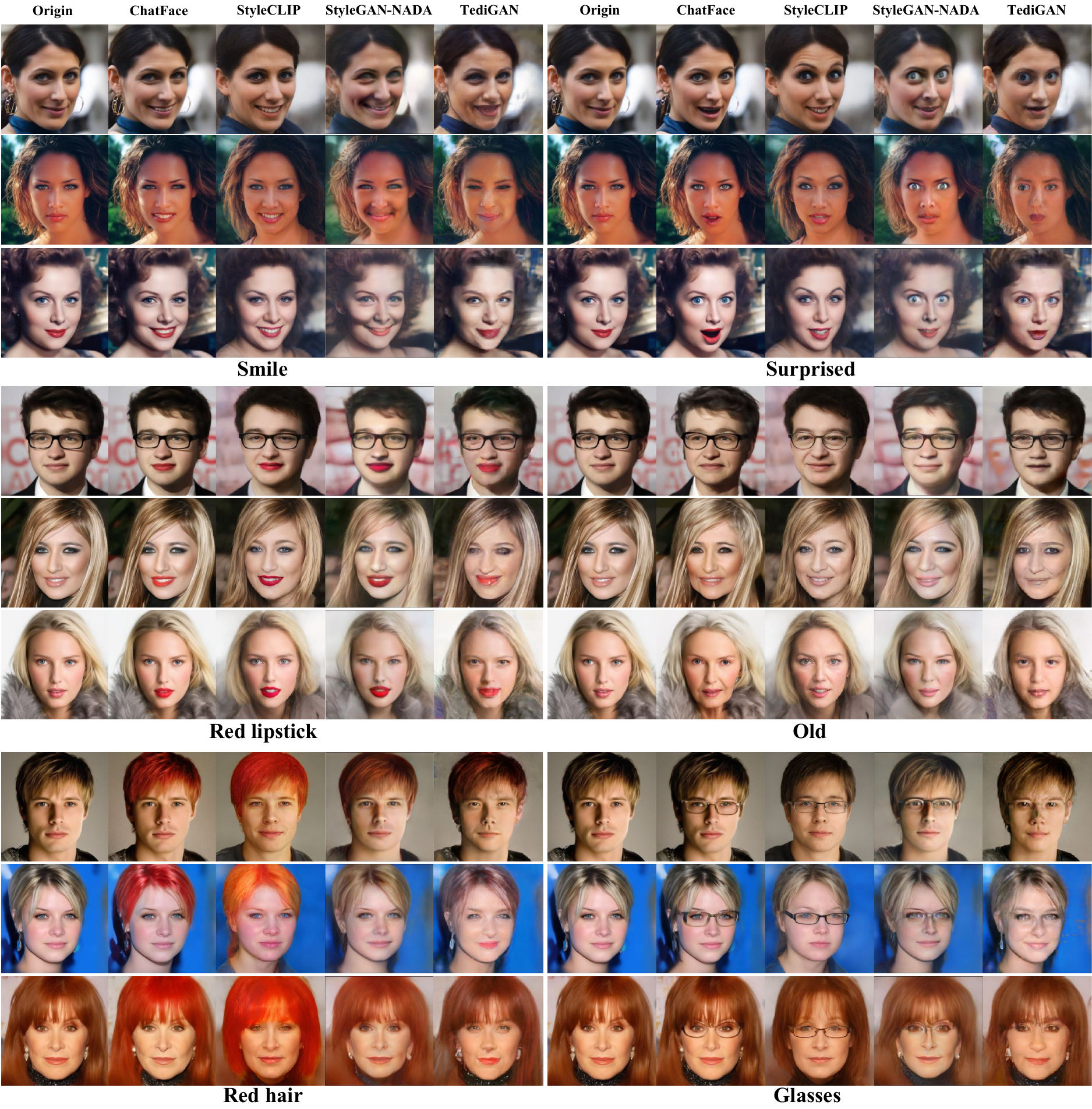}
 \caption{Comparison with more GAN inversion-based manipulation.}
 \label{fig:app_compare}
\end{figure}

\textbf{Comparison with GAN-based methods.}
In Figure\ref{fig:app_compare}, we present a comparison between ChatFace and GAN-based image manipulation methods. The results demonstrate that despite using state-of-the-art inversion techniques, these GAN-based methods still struggle to faithfully preserve the undesired semantics of the input image, such as background and accessories.

\begin{figure}[htp]
 \centering
 \includegraphics[width=1.0\columnwidth]{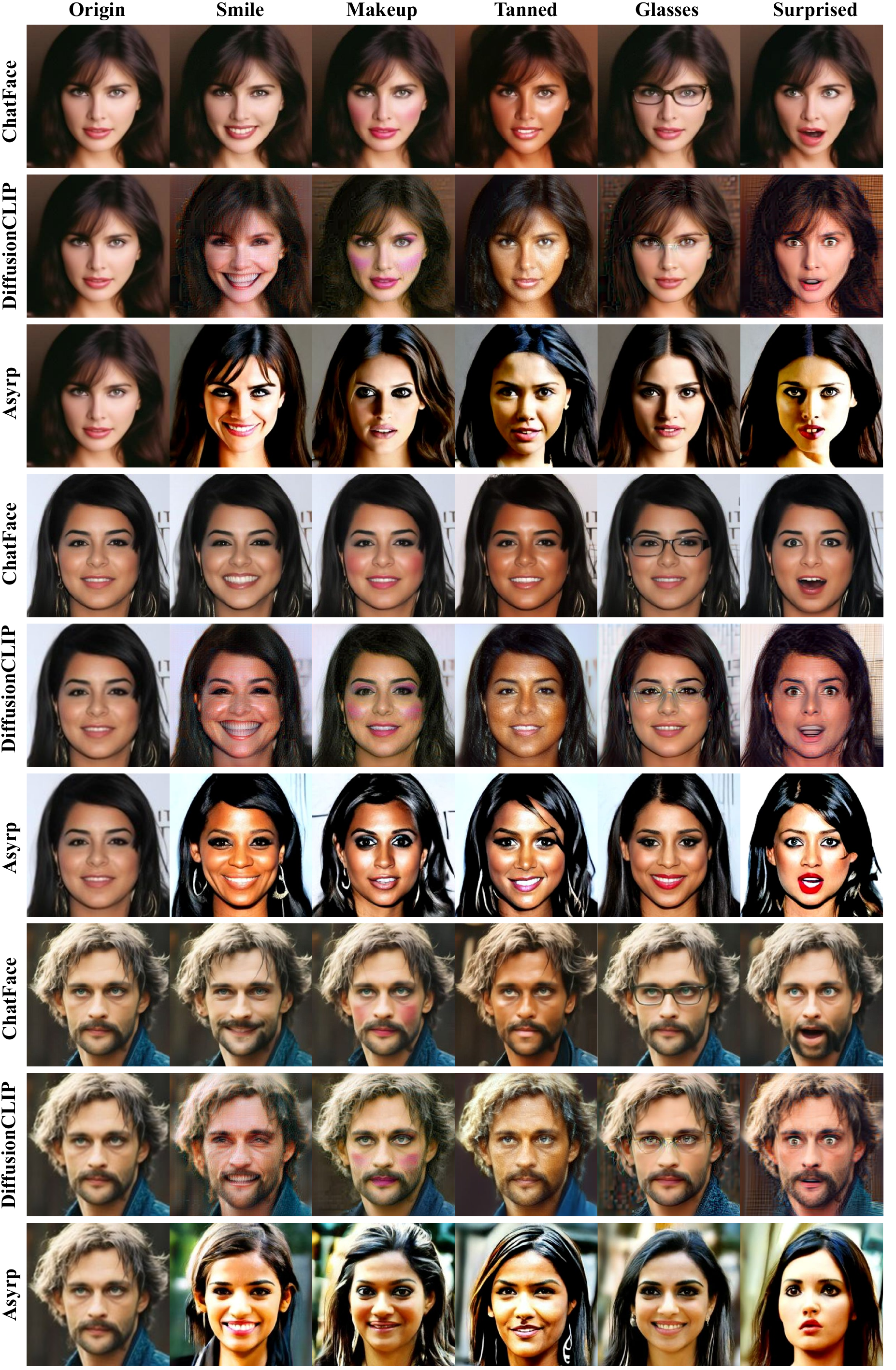}
 \caption{More comparison results with diffusion-based methods: DiffusionCLIP\cite{kim2022diffusionclip} and Asyrp\cite{kwon2022diffusion}.}
 \label{fig:app_compare_diff}
\end{figure}

\textbf{Comparison with diffusion-based methods.}
We further compared the diffusion model-based image manipulation methods, as shown in Figure\ref{fig:app_compare_diff}. The results demonstrate that our ChatFace is capable of more accurately manipulating the semantic aspects of facial images while preserving the details of the original image.

\section{Case Study}\label{sec:appB}
\begin{figure}[htp]
 \centering
 \includegraphics[width=1.0\columnwidth]{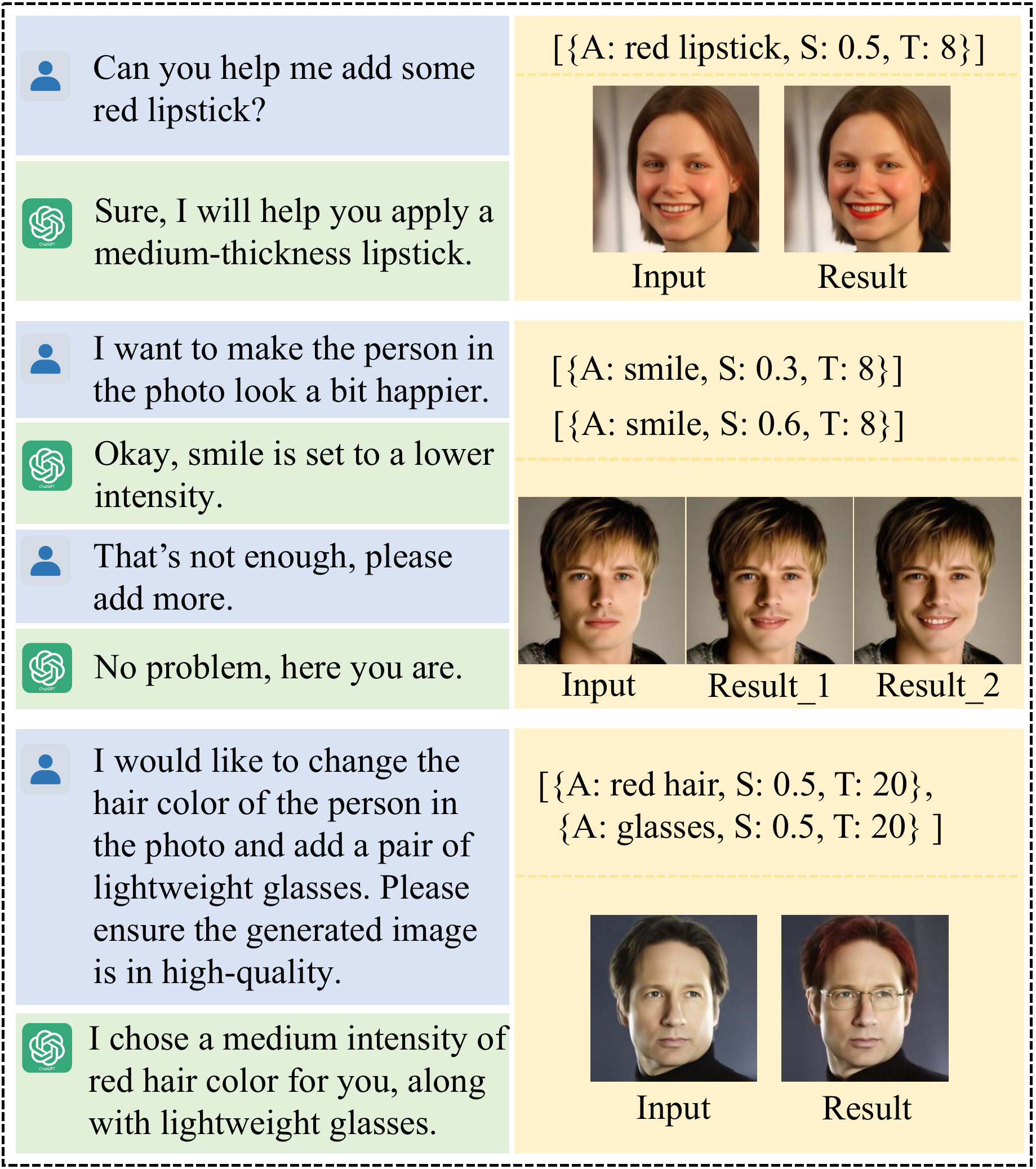}
 \caption{Case study on simple and complex editing tasks.}
 \label{fig:app_case}
\end{figure}

ChatFace is a multimodal system that combines large language models with the diffusion model's capacity to manipulate the real face images within the semantic latent space of the diffusion model through interactive dialogue. We tested ChatFace on a wide range of multimodal image editing tasks, and selected cases are shown in Figure\ref{fig:app_case}. ChatFace can solve multiple tasks such as single-facial attribute editing, interactive editing with strength control, and complex multi-attribute editing. It also supports user-defined expectations for image quality. Higher quality images require more diffusion generation time steps.

\section{Ablition Study and Hyperparameter}\label{sec:appC}
\subsection{Effect of Stable Manipulation Strategy}
The Stable Manipulation Strategy (SMS) achieves more reliable semantic manipulation by matching the semantic dimensions of the diffusion model in the temporal domain. To demonstrate the necessity of SMS (Stable Manipulation Strategy), we conducted quantitative comparative experiments, and the results are presented in Table\ref{tab:app_SMS}.

\subsection{Dependency on Generation Time Steps $T$}
\begin{figure}[htp]
 \centering
 \includegraphics[width=1.0\columnwidth]{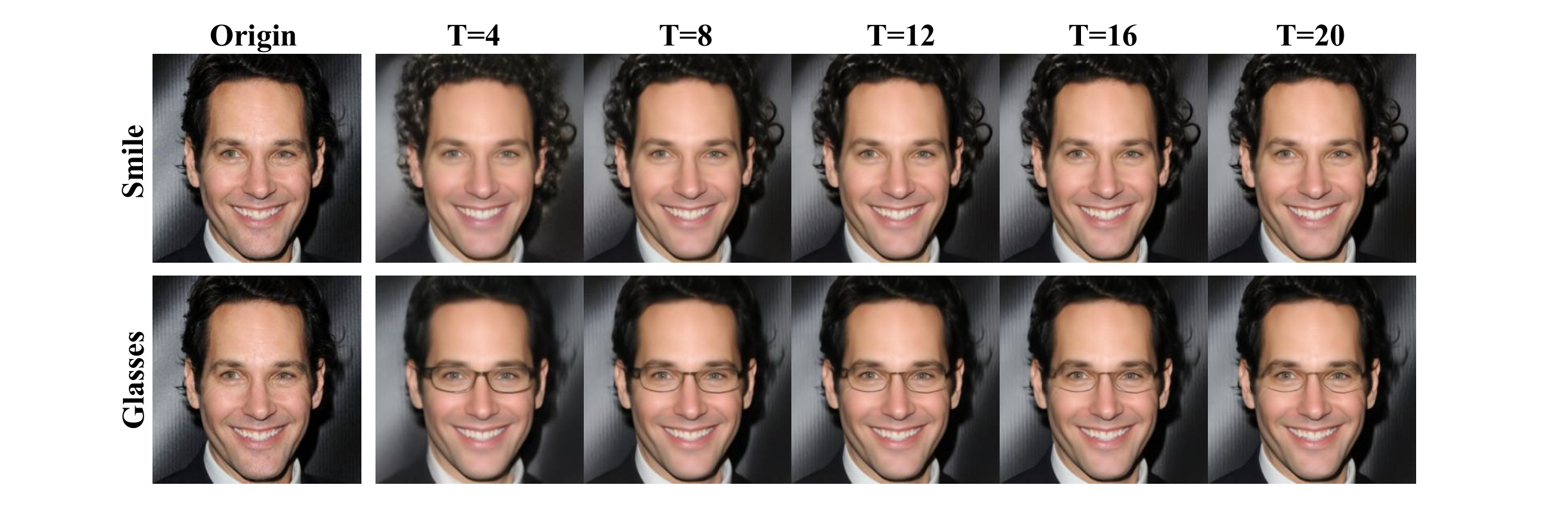}
 \caption{The effect of the number of diffusion generation sample time steps. We analyze $T=$ 4, 8, 12, 16 and 20 and choose to use 8 time steps.}
 \label{fig:app_ablation_t}
\end{figure}

In the configuration of ChatFace, we use $T=8$ as the default for the generation sampling steps unless explicitly specified by the user. Figure\ref{fig:app_ablation_t} illustrates the reason for this setting. By observing the results, it can be noticed that when the number of sampling steps is smaller, ChatFace with SMS produces a higher editing strength result but loses high-frequency information from the input image, such as background patterns. As the number of sampling steps increases, the detailed information of the image is more fully restored, but it also requires a longer time and weakens the editing strength. Therefore, we strike a balance between the consistency of real image editing and the smoothness of the interactive experience.

\renewcommand{\arraystretch}{1.2}
\begin{table}[h]
    \caption{The details of the prompt design and semantic activation in ChatFace.}
    \label{tab:demanstration}
    \centering
    \small
    \begin{tabular}{|m{0.4cm}|m{3.5cm}m{8.5cm}|}
        \toprule
        \multirow{23}{*}{\rotatebox{90}{Editing Intention Understanding}} & \multicolumn{2}{C{12cm}|}{Prompt} \\ [2pt]
       \cline{2-3}
        & \multicolumn{2}{L{12.35cm}|}{\#1 Editing Intention Understanding Stage - You are an expert linguist. You need to summarize various situations based on existing knowledge and then select a reasonable solution. You need to parse user input to several tasks: \ul{[\{"task": task, "id": mapper\_id, "args": \{"attribute": attribute, "strength": strength\_score, "time\_steps": sample\_time\_steps\}\}]}. The task must be selected from the following options: \{\{Available Mapper List\}\}. You need to learn how to identify the subject, the descriptive words of the editing strength, and the descriptive words of image clarity from a sentence, and convert the latter two into floating-point numbers between 0 and 1, and integers between 8 and 50, respectively. The higher the numerical value, the stronger the degree. You need to read and understand the following examples: \{\{Demonstrations\}. From the chat logs, you can find the path of the user-mentioned resources for your task planning.}\\
       \cline{2-3}
        & \multicolumn{2}{C{12.35cm}|}{Demonstrations}\\
       \cline{2-3}
        & Can you help me add some smiles to the people in the photo? & [\ul{\{"task": smile, "id": 0, "args": \{attribute": smile, "strength": 0.5, "time\_steps": 8\}\}}]\\
        & I would like to make this face look younger and the skin a bit lighter. & [\ul{\{"task": young, "id": 1, "args": \{attribute": young, "strength": 0.5, "time\_steps": 8\}\}}, \ul{\{"task": pale, "id": 2, "args": \{attribute": pale, "strength": 0.2, "time\_steps": 8\}\}}]\\
        & I would like to try curly hair and also add a deep red lipstick.  & [\ul{\{"task": curly hair, "id": 3, "args": \{attribute": curly hair, "strength": 0.5, "time\_steps": 8}, \ul{\{"task": red lipstick, "id": 4, "args": \{attribute": red lipstick, "strength": 0.9, "time\_steps": 8\}\}}]\\
        & Please help me generate a clear photo of me wearing glasses and with light makeup. & [\ul{\{"task": glasses, "id": 5, "args": \{attribute": glasses, "strength": 0.5, "time\_steps": 20\}\}}, \ul{\{"task": makeup, "id": 6, "args": \{attribute": makeup, "strength": 0.2, "time\_steps": 20\}\}}]\\
        \midrule
        \multirow{11}{*}{\rotatebox{90}{Semantic Activation\quad\quad\quad}}
       & \multicolumn{2}{L{12.35cm}|}{\#2 Semantic Activation Stage - The primary aim of this stage is to establish a successful alignment between the parsed requests and the editing offset in diffusion semantic latent space. To accomplish this, we segment the mapping network into distinct words that are likely to occur and apply regular expressions to standardize formats, including capitalization and underscores. Consequently, the mapper\_id will correspond to a list that potentially contains the relevant matches for that mapping network. Thus significantly enhancing the overall alignment and performance of the system.} \\
        \cline{2-3}

       & \multicolumn{2}{C{12.35cm}|}{Available Mapper List}\\
       \cline{2-3}
        & mapper\_id & mapper\\
        & 0 & [smile, smiling, happy]\\
        & 1 & [young, without wrinkle]\\
        & 2 & [pale, white, whiter]\\
        & 3 & [curly hair, hair curly]\\
        & 4 & [red lipstick, red lip stick, lipstick red, lip stick red]\\
        & 5 & [glasses]\\
        & 6 & [makeup]\\

       
       \bottomrule
    \end{tabular}

    \label{tab:prompt}
\end{table}

\begin{table}[]
\caption{Quantitative ablation analysis results.}  
\label{tab:app_SMS}
\centering
    \setlength\tabcolsep{2.0mm}
    \renewcommand\arraystretch{1.1}
    {
    \scalebox{1.0}{
        \begin{tabular}{c ccc}
        \toprule[1pt]
        \textbf{}              & \multicolumn{3}{c}{Editing Performance}                          \\
        \cmidrule[1pt](lr){2-4}
        & $S_{dir}\uparrow$    & SC$\uparrow$   & ID$\uparrow$ \\
        \midrule
        w/o SMS         & 0.18                & 88.3\%              & 0.83              \\
        Ours      & \textbf{0.21}                & \textbf{89.7\%}              &  \textbf{0.84}        \\
        \bottomrule[1pt]
        \end{tabular}}
    }

\end{table}

\end{appendices}

\end{document}